\documentclass{article}

\usepackage[final]{neurips_data_2022}

\usepackage[dvipsnames]{xcolor}

\usepackage{amsmath,amsfonts,bm}

\def\eqref#1{equation~\ref{#1}}

\def\1{\bm{1}}

\DeclareMathAlphabet{\mathsfit}{\encodingdefault}{\sfdefault}{m}{sl}
\SetMathAlphabet{\mathsfit}{bold}{\encodingdefault}{\sfdefault}{bx}{n}

\usepackage{color,xcolor}
\usepackage{epsfig}
\usepackage{graphicx}

\usepackage{adjustbox}
\usepackage{array}
\usepackage{booktabs}
\usepackage{colortbl}
\usepackage{wrapfig}
\usepackage{hhline}
\usepackage{multirow}
\usepackage{subcaption} 
\usepackage{floatflt}

\usepackage{amsmath,amsfonts,amssymb}
\usepackage{mathtools}  
\usepackage{bm}
\usepackage{nicefrac}
\usepackage{microtype}

\usepackage{changepage}
\usepackage{extramarks}
\usepackage{fancyhdr}
\usepackage{lastpage}
\usepackage{setspace}
\usepackage{soul}
\usepackage{xspace}

\usepackage[pagebackref=true,breaklinks=true,colorlinks,citecolor=gray]{hyperref}

\usepackage{url}

\usepackage{enumerate}
\usepackage{todonotes} 
\usepackage{enumitem}  

\usepackage{titlesec}

\usepackage{makecell}

\usepackage{pifont} 
\usepackage{subcaption}

\usepackage{algorithm,algpseudocode}

\usepackage{amsthm}
\newcolumntype{L}[1]{>{\raggedright\let\newline\\\arraybackslash\hspace{0pt}}m{#1}}
\newcolumntype{C}[1]{>{\centering\let\newline\\\arraybackslash\hspace{0pt}}m{#1}}
\newcolumntype{R}[1]{>{\raggedleft\let\newline\\\arraybackslash\hspace{0pt}}m{#1}}

\newcommand{\ignore}[1]{}

\makeatletter
\DeclareRobustCommand\onedot{\futurelet\@let@token\@onedot}
\def\@onedot{\ifx\@let@token.\else.\null\fi\xspace}

\def\eg{e.g\onedot}

\makeatother

\definecolor{MyDarkBlue}{rgb}{0,0.08,1}
\definecolor{MyDarkGreen}{rgb}{0.02,0.6,0.02}
\definecolor{MyDarkRed}{rgb}{0.8,0.02,0.02}
\definecolor{MyDarkOrange}{rgb}{0.40,0.2,0.02}
\definecolor{MyPurple}{RGB}{111,0,255}
\definecolor{MyRed}{rgb}{1.0,0.0,0.0}
\definecolor{MyGold}{rgb}{0.75,0.6,0.12}
\definecolor{MyDarkgray}{rgb}{0.66, 0.66, 0.66}

\newcommand{\model}{Geoclidean\xspace}
\newcommand{\sdiff}{\textit{Close}\xspace}
\newcommand{\ldiff}{\textit{Far}\xspace}
\newcommand{\lfeat}{low\xspace}
\newcommand{\hfeat}{high\xspace}
\newcommand{\elements}{Geoclidean-Elements\xspace}
\newcommand{\constraints}{Geoclidean-Constraints\xspace}

\usepackage[utf8]{inputenc} 
\usepackage[T1]{fontenc}    
\usepackage{hyperref}       
\usepackage{url}            
\usepackage{booktabs}       
\usepackage{amsfonts}       
\usepackage{nicefrac}       
\usepackage{microtype}      
\usepackage{xcolor}         
\usepackage{float}

\title{\model: Few-Shot Generalization \\in Euclidean Geometry}

\author{
  Joy Hsu \\
  Computer Science \\
  Stanford University\\
  \texttt{joycj@stanford.edu} \\
  \And
  Jiajun Wu \\
  Computer Science \\
  Stanford University\\
  \texttt{jiajunwu@cs.stanford.edu} \\
  \And
Noah D.~Goodman \\
  Psychology and Computer Science \\
  Stanford University\\
  \texttt{ngoodman@stanford.edu} \\
}

\begin{document}

\maketitle

\begin{abstract}
Euclidean geometry is among the earliest forms of mathematical thinking. While the geometric primitives underlying its constructions, such as perfect lines and circles, do not often occur in the natural world, humans rarely struggle to perceive and reason with them. Will computer vision models trained on natural images show the same sensitivity to Euclidean geometry? Here we explore these questions by studying few-shot generalization in the universe of Euclidean geometry constructions. We introduce \emph{\model}, a domain-specific language for Euclidean geometry, and use it to generate two datasets of geometric concept learning tasks for benchmarking generalization judgements of humans and machines. We find that humans are indeed sensitive to Euclidean geometry and generalize strongly from a few visual examples of a geometric concept. In contrast, low-level and high-level visual features from standard computer vision models pretrained on natural images do not support correct generalization. Thus \model represents a novel few-shot generalization benchmark for geometric concept learning, where the performance of humans and of AI models diverge. The \model framework and dataset are publicly available for download.\footnote{The Geoclidean framework can be found at \url{https://github.com/joyhsu0504/geoclidean_framework}.}
\footnote{Datasets can be found at \url{https://downloads.cs.stanford.edu/viscam/Geoclidean/geoclidean.zip}.}
\end{abstract}


\section{Introduction}

The built world we inhabit is constructed from geometric principles. Yet geometric primitives such as perfect lines and circles, which are the foundations of human-made creations, are uncommon in the natural world. Whether for efficiency or for visual aesthetics, whether innate or learned, humans are sensitive to geometric forms and relations. This natural understanding of geometry enables a plethora of applied skills such as design, construction, and visual reasoning; it also scaffolds the development of rigorous mathematical thinking, historically and in modern education. Thus, understanding the visually-grounded geometric universe is an important desideratum for machine vision systems.

Ancient Greek philosophers were amongst the earliest to formalize geometric notions, culminating in Euclid’s geometry in the 4th century BC. With a compass and straight edge, Euclid’s axioms can construct a geometric world that reflects an idealized, or Platonic, physical reality. We hypothesize that  Euclidean constructions are intrinsic to human visual reasoning. We thus build a library to define and render such concepts, allowing systematic exploration of geometric generalization. In this paper, we present \model, a domain-specific language (DSL) for describing Euclidean primitives and construction rules.
Sets of construction rules define concepts that can then be realized into infinitely many rendered images capturing the same abstract geometric model. 
\begin{figure}
  \begin{small}
  \begin{center}
    \includegraphics[width=1.0\textwidth]{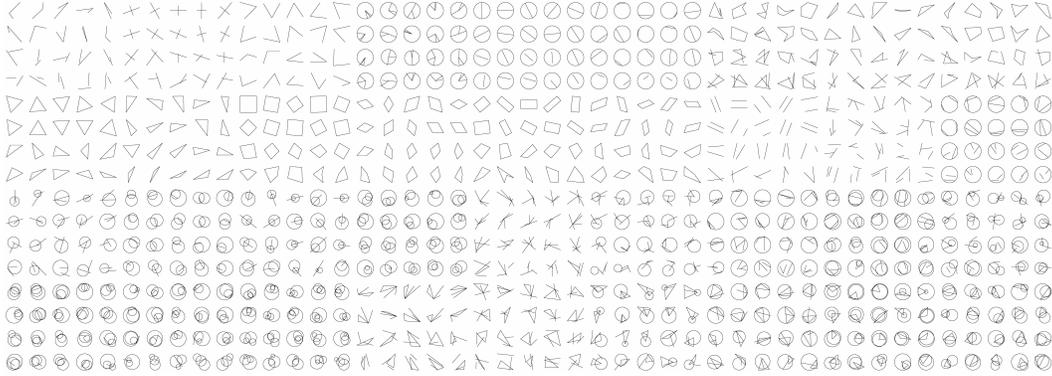}
  \end{center}
  \end{small}
  \caption{Rendered realizations of Euclidean geometry concepts from the \model datasets.}
\label{fig:pull}
\end{figure}

Based on \model, we introduce two datasets to study few-shot generalization to novel rendered realizations of Euclidean geometry concepts (See Figure~\ref{fig:pull}). To succeed in solving these tasks, one must understand the underlying geometric concept of a set of rendered images. 
The first dataset, \elements, covers mathematical definitions from the first book of Euclid's Elements \citep{simson1838elements}.
The second dataset, \constraints, simplifies and more systematically explores possible relationships between primitives.
We publicly release both datasets, as well as the dataset generation library based on \model.

We report findings on the \model few-shot generalization tasks from human experiments, as well as from evaluation on low-level and high-level visual features from ImageNet-pretrained VGG16 \citep{deng2009imagenet, simonyan2014very}, ResNet50 \citep{he2016deep}, InceptionV3 \citep{szegedy2016rethinking}, and Vision Transformer \citep{dosovitskiy2020image}. We show that humans significantly outperform pretrained vision models, generalizing in a way that is highly consistent with abstract target concepts. ImageNet-pretrained vision models are not as sensitive to intrinsic geometry and do not as effectively encode these geometric abstractions. Our benchmarking process illustrates this gap between humans and models, establishing \model as an interesting and challenging generalization task for visual representation and geometric concept learning. 

\section{Foundations of \model} 
In this section, we give an overview of the \model DSL, which builds on foundations of Euclid’s axioms for constructing with a compass and straightedge. We present a Python library that renders Euclidean geometry concepts described from our DSL into images. We first describe the Euclidean geometry universe in Section~\ref{euclidean_geometric_universe}, and then introduce the \model language and construction rules in Section~\ref{model_concept_language}. Finally, we show geometric concept realizations into rendered images in Section~\ref{concept_realizations}. 

\subsection{Euclidean Geometry Universe}
\label{euclidean_geometric_universe}
There exist numerous systems of geometry, each with its own logical system. Each logic can be described in a formal language, each formal language describes construction rules, and each set of construction rules describe \textit{concepts} in the geometric universe. Geometric concepts can be realized and rendered into images. 
Euclidean geometry was among the earliest formalized, first described by Euclid in Elements \citep{simson1838elements}. The first book of Elements details plane geometry, laying the foundation for basic properties and propositions of geometric objects. Importantly, Euclid’s geometry is constructive---objects only exist if they can be created with a compass and straightedge. Euclidean constructions are \textit{abstract} models of geometric objects, specified without the use of coordinates and without concrete realizations into images. Hence, changes to the image rendering, such as alterations to size and rotation, do not change an object’s intrinsic Euclidean geometry. 

Euclid’s axioms build the foundations of this geometric universe via 1) draw a straight line from any point to any point, 2) describe a circle with any centre and distance. The Euclidean geometry universe is determined by these base geometric \textit{primitives}, and the \textit{constraints} that parameterize object relationships. There are other universes that contain different logic systems, from Descartes' analytic geometry, to Euler's affine geometry, to Einstein's special relativity. 
Because of its simplicity and abstraction we are particularly interested in the universe of Euclidean geometry -- do humans and machines find these constructions natural? We later explore whether they spontaneously generalize image classes according to Euclidean rules.






\subsection{Domain-Specific Language: \model}
\label{model_concept_language}

We create a domain-specific language (DSL), \model, defining Euclidean constructions that arise from the mathematics of Euclidean geometry. \model allows us to define construction rules for objects and their relations to each other, encompassing concepts in the Euclidean geometry universe. It includes three simple primitives. The first is a \textbf{point}, parameterized by constraints, if any, to an object or a set of objects previously defined. The point is defined without specific coordinates, and only when \textit{realized}, would be assigned coordinate values $x$ and $y$. The second is a \textbf{line}, which, following the first axiom, is parameterized by two points, representing the beginning and end. The third is a \textbf{circle} which, following the second axiom, is defined by a center point and an edge point. These primitives represent the compass and straightedge constructions that Euclid introduced. Lines and circles are defined by points, while points
can be constrained to previously built objects. In this way primitives are sequentially defined to form a \textit{concept}.


\begin{table*}[htb!] 
\caption{The \model DSL for building Euclidean geometry concepts. We assume a pool of variable names for points and objects (\texttt{point\_name}, \texttt{object\_name}). As a shorthand for point creation followed by reference, we later use e.g. \texttt{Line(p1(),p2())} to represent \texttt{p1 = Point(); \dots~Line(p1,p2)}. The marker * indicates that the object will not be visible in the final rendering.}
\begin{center}
\begin{small}
\begin{sc}
\begin{tabular}{lll}
      \toprule
      Concept & $\rightarrow$ & Statement; Concept \\
      Statement & $\rightarrow$ & \texttt{object\_name}$^\text{Visibility}$ = Object(\texttt{point\_name}, \texttt{point\_name}) | \\
      && \texttt{point\_name} = \textsf{point}(Constraints) \\
      Visibility & $\rightarrow$ & [] | * \\
      Object & $\rightarrow$ & \textsf{line} | \textsf{circle} \\
      Constraints & $\rightarrow$ & [] | [\texttt{object\_name}] | [\texttt{object\_name}, \texttt{object\_name}] \\
      \bottomrule
  \end{tabular}
\end{sc}
\end{small}
\end{center}
\label{table:dsl}
\end{table*}

\model's syntax is defined in Table~\ref{table:dsl}. We can initialize a new point, line, or circle via the constructors \texttt{Point}, \texttt{Line}, \texttt{Circle}, assigning them to a named variable.
For points we generally use a shorthand to define and use the variable inline.
For example, \texttt{p1()} is a free point that is unconstrained to any objects, and can be realized anywhere in the image, \texttt{p2(circle1)} is a partially constrained point that lies on the object \texttt{circle1}, and \texttt{p3(line1, line2)} is a fully constrained point that lives in the intersection of \texttt{line1} and \texttt{line2}. A point can be reused by referring to its name. The line \texttt{Line(p1, p2)} is parameterized by two end points, and the circle \texttt{Circle(p1, p2)} is parameterized by points at the center and edge. 
Not indicated in Table~\ref{table:dsl} is the semantic constraint that variable names must be defined before being used within constructors.
Visibility of rendering is denoted for each object, with * indicating that the object is \textit{not} visible in the final rendering, as some objects in Euclidean geometry are used solely as helper constructions for other core objects in the concept. Each construction rule is a geometric object, and a sequence of construction rules define a Euclidean geometry concept.

\subsection{Concept Realization}
\label{concept_realizations}

\model implements this Euclidean language and renders realizations of concepts from the language into images. For each line or circle object, we realize its rendering based on the point parameters required. During rendering, each point parameter is given randomly sampled real-valued coordinates bound by its constraints. If the point exists already, \model reuses the past component; if it is a free point without constraints, \model randomly samples values for $x$ and $y$; if constrained, \model randomly samples a point that lies constrained on the object or the intersection of a set of objects. It is possible for intersection sets to be empty, and we reject sample realizations until finding a satisfying realization for all points. \model creates the objects sequentially and renders them into an image, if visible. 

\vspace{0em} 
\begin{minipage}[m]{0.65\textwidth}
\begin{algorithm}[H] 
\caption{Construction rules for the equilateral triangle concept, with colored steps corresponding to the rendered realization in Figure~\ref{fig:construction}. The two circles \texttt{c1} and \texttt{c2} are not rendered in the final image as denoted by *.} 
\label{alg:syntax}
\begin{algorithmic}[1]
\item[]
\State \texttt{l1 = Line(p1(), p2())} \Comment{\textcolor{BrickRed}{Red} line}
\State \texttt{c1*= Circle(p1(), p2())} \Comment{\textcolor{Dandelion}{Light orange} circle}
\State \texttt{c2*= Circle(p2(), p1())} \Comment{\textcolor{GreenYellow}{Light yellow} circle}
\State \texttt{l2 = Line(p1(), p3(c1, c2))} \Comment{\textcolor{OliveGreen}{Green} line}
\State \texttt{l3 = Line(p2(), p3(c1, c2))} \Comment{\textcolor{BlueViolet}{Blue} line}
\end{algorithmic}
\end{algorithm}
\end{minipage}
\hfill
\begin{minipage}[m]{0.33\textwidth}
\begin{figure}[H]
  \begin{small}
  \begin{center}
    \includegraphics[width=0.9\textwidth]{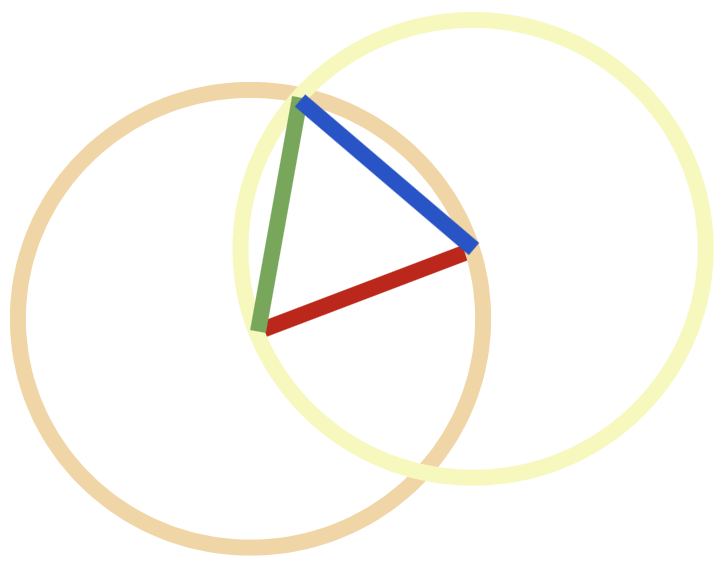}
  \end{center}
  \end{small}
  \caption{Rendered realization of the equilateral triangle concept.} 
\label{fig:construction}
\end{figure}
\end{minipage}

We see in Figure~\ref{fig:construction} an example of the \model language describing the equilateral triangle concept realized into an image. The construction rules in Algorithm~\ref{alg:syntax} are created step by step; we color each object for clarity. The first construction rule creates a red line between two unconstrained free points, \texttt{p1()} and \texttt{p2()}; when realized with sampled real-valued points, this line can be anywhere in the image with any length and rotation. The second rule creates an invisible orange circle with the ends of the first line as its center and edge point. The third rule creates another invisible yellow circle with the center and edge point flipped, forming intersecting helper circles with the same radius. Then, the fourth rule creates a green line between \texttt{p1} and a new point \texttt{p3(c1, c2)}, which is a point constrained to the intersection of the previously created orange and yellow circles (the realization randomly chooses one of the two intersection points). The last rule creates the final blue line between \texttt{p2} and \texttt{p3}, completing the constructed triangle and enforcing all sides to be of equal length. This concept is that of the \textit{equilateral triangle}, and we see that the invisible helper circle objects serve as essential constraints to the final rendering.

Importantly, the construction rules do not specify any coordinates, and our \model framework creates the coordinates upon realization of the concept into an image. Hence, the rendered equilateral triangle can be of any size and orientation, while always respecting the underlying geometric concept. In \model, every random realization of this concept creates equilateral triangles, as represented in Euclid’s geometric universe. The \model framework allows us to create rendered image datasets that follow the specified concept language. 

\section{\model Task and Datasets}
Do the realizations of Euclidean constructions form natural categories for humans? For computer vision models?
To study these questions we introduce a few-shot generalization task based on \model and two image datasets that realize $37$ Euclidean geometry concepts. 

\paragraph{Task.}
\label{model_task}
We explore few-shot generalization from positive examples of a target concept; to the extent that participants generalize to other realizations of the target concept, and not realizations of more general concepts, we conclude the concept is an intuitive kind. 

\begin{figure}
  \begin{small}
  \begin{center}
    \includegraphics[width=1.0\textwidth]{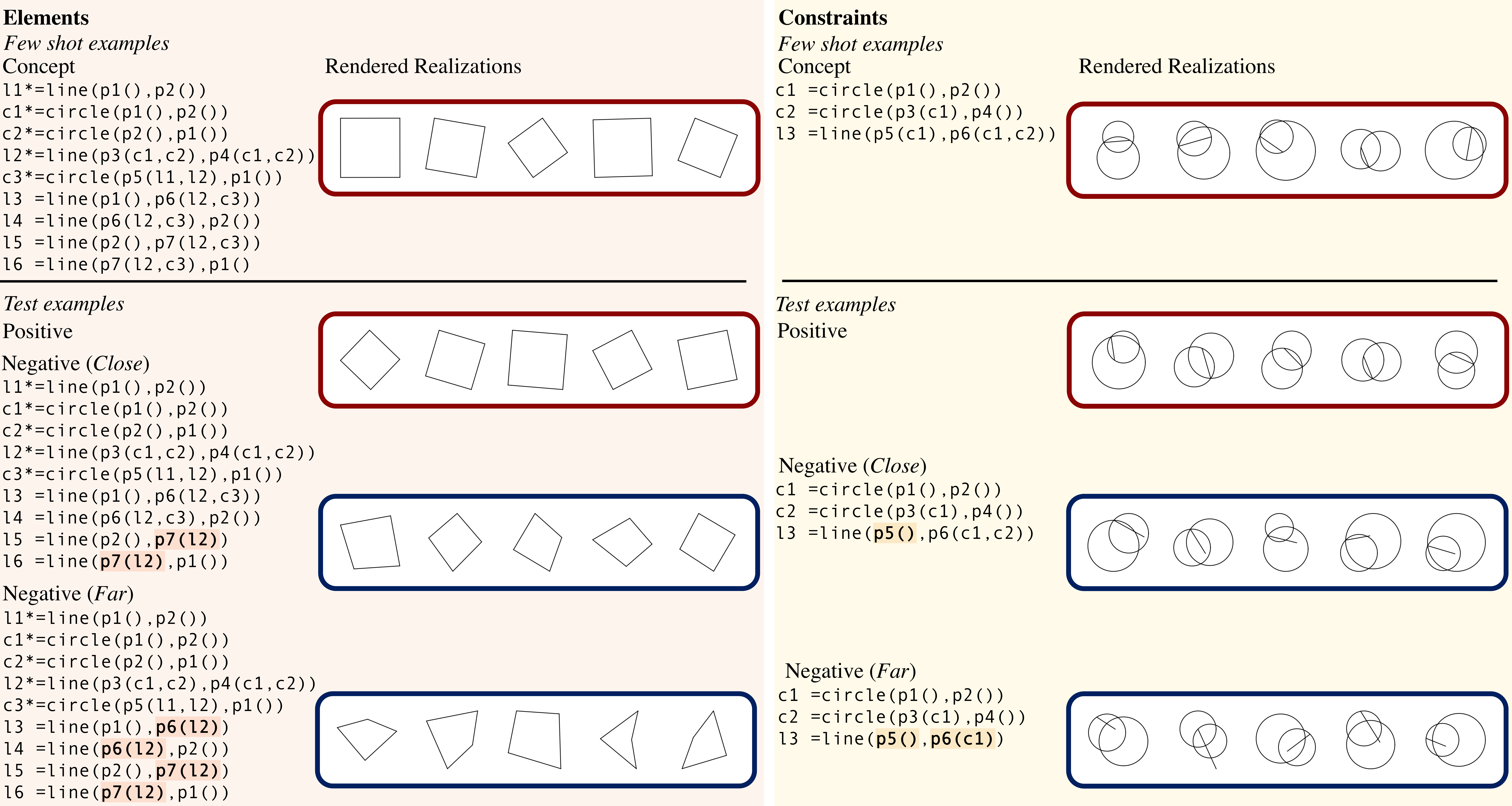}
  \end{center}
  \end{small}
  \caption{Examples of \elements and \constraints tasks. Each task consists of few-shot reference examples as well as test examples. Few-shot examples and positive test examples derive from the same concept (contained in red boxes), while negative test examples derive from a related but different concept (contained in blue boxes); \sdiff test examples differ by fewer point differences in the construction rules than \ldiff, seen bolded in the \model language.}
\label{fig:data_examples}
\end{figure}

For each concept, the task includes five reference examples of the target concept and a test set of 15 images. Among the 15 images, there are five positive examples and two sets of five negative examples. Here, positive examples in the test set are realizations of the target concept, as are the reference examples, and negative examples are realizations of related but different concepts (which are not realizations of the target concept). 
The goal is to correctly categorize positive examples as positive and negative examples as negative in the test set. The ten negative examples are divided into five \sdiff and five \ldiff examples, where the negative examples in \sdiff are from a closely related concept with a fewer number of constraint differences from the target concept, and negative examples in \ldiff are from a further, less related concept with a larger number of constraint differences. These constraint differences consist of altering a point to have fewer constraints compared to that point in the target concept (yielding a more general and less specified geometric concept). See Figure~\ref{fig:data_examples} for examples, with the top representing reference examples, and the bottom representing the test set.
Note that, because we are interested in intrinsic sensitivity of visual representations to geometric concepts, we are not introducing a meta-learning task: there are no few-shot generalization sets intended for model training.

We now introduce the two datasets we created based on \model for the generalization task. The first dataset, \elements, includes the tests of $17$ concepts derived from the first book of Euclid's Elements; the second dataset, \constraints, includes the tests of $20$ concepts based on constraints defining relationships between \model primitives. See Figure~\ref{fig:data_examples} for examples from both splits. 

\paragraph{\elements.}
The \elements dataset is derived from definitions in the first book of Euclid's Elements, which focuses on plane geometry. \elements includes $17$ target concepts, which, along with \model primitives, covers definitions in the first book of Elements. These concepts require complex construction rules with helper objects that are not visible in the final renderings. Realizations from \elements test sensitivity to the exactness of Euclidean constructions without explicit visual constraint differences.

The concepts in \elements include angle (Book I definition IX), perpendicular bisector (def X), angle bisector (def X), sixty degree angle (def XI and XII), radii (def XV), diameter (def XVII and XVIII), segment (def XIX), rectilinear (def XX and XXIII), triangle (def XXI), quadrilateral (def XXII and XXXIV), equilateral triangle (def XXIV, XXV, XXVI in \sdiff and \ldiff), right angled triangle (def XXVII, XXVIII, XXIX in \sdiff and \ldiff), square (def XXX), rhombus (def XXXI), oblong (def XXXII), rhomboid (def XXXIII), and parallel lines (def XXXV). The rest of the definitions are descriptions of \model primitives (e.g.~points, lines).

\paragraph{\constraints.}
The \constraints dataset consists of $20$ concepts, created from permutations of line and circle construction rules with various constraints describing the relationship between objects. This dataset focuses on explicit constraints between geometric objects. We denote the objects as the following---lines as \texttt{L}, circles as \texttt{C}, and triangles (constructed from three lines) as \texttt{T}. Tasks include three, four, and five object variants, each with specific ordering; the different ordering of objects is significant, as constraints may only depend on previously defined objects. Each concept is defined by object ordering as well as constraints describing the relationships between them; the full set of construction rules for each concept is released with the dataset.

The three object concepts are \texttt{[LLL, CLL, LLC, CCL, LCC, CCC]}, the four object concepts are \texttt{[LLLL, LLLC, CLLL, CLCL, LLCC, CCCL, CLCC, CCCC]}, and the five object concepts are \texttt{[TLL, LLT, TCL, CLT, TCC, CTT]}. These $20$ concepts test the few-shot generalization capability in constrained Euclidean geometry concepts.

\section{Findings}
We present our findings on the \model dataset in benchmarking human performance (Section~\ref{human_performance}) and pretrained vision models' capabilities (Section~\ref{model_benchmarks}). We show that humans are indeed sensitive to Euclidean geometry concepts, generalizing strongly from five examples across the $37$ concepts. 
This establishes \model as an interesting task for evaluating the human-like visual competencies of machine vision. Indeed, we find that state-of-the-art pretrained visual representations perform poorly on this few-shot generalization task. 

\subsection{Human Performance}
\label{human_performance}
We collected human judgements for the \model few-shot concept learning task. We recruited $30$ participants for each concept using the Prolific crowd-sourcing platform \citep{palan2018prolific}. 
As mentioned above, participants are given five example realizations of the target concept and $15$ test images including five positive examples, five \sdiff negative examples, and five \ldiff negative examples. Each of the test questions states: “Each of the five images shown is a ‘wug’. Is the image below a ‘wug’ or not a ‘wug’?”, where ‘wug’ is a random made-up word for each task. The order of tasks and realizations is randomized for each participant to remove order effects. The experiment interface was implemented on Qualtrics, with details described in the Appendix. 

\begin{wrapfigure}{r}{0.7\textwidth}
  \begin{small}
  \begin{center}
    \includegraphics[width=0.7\textwidth]{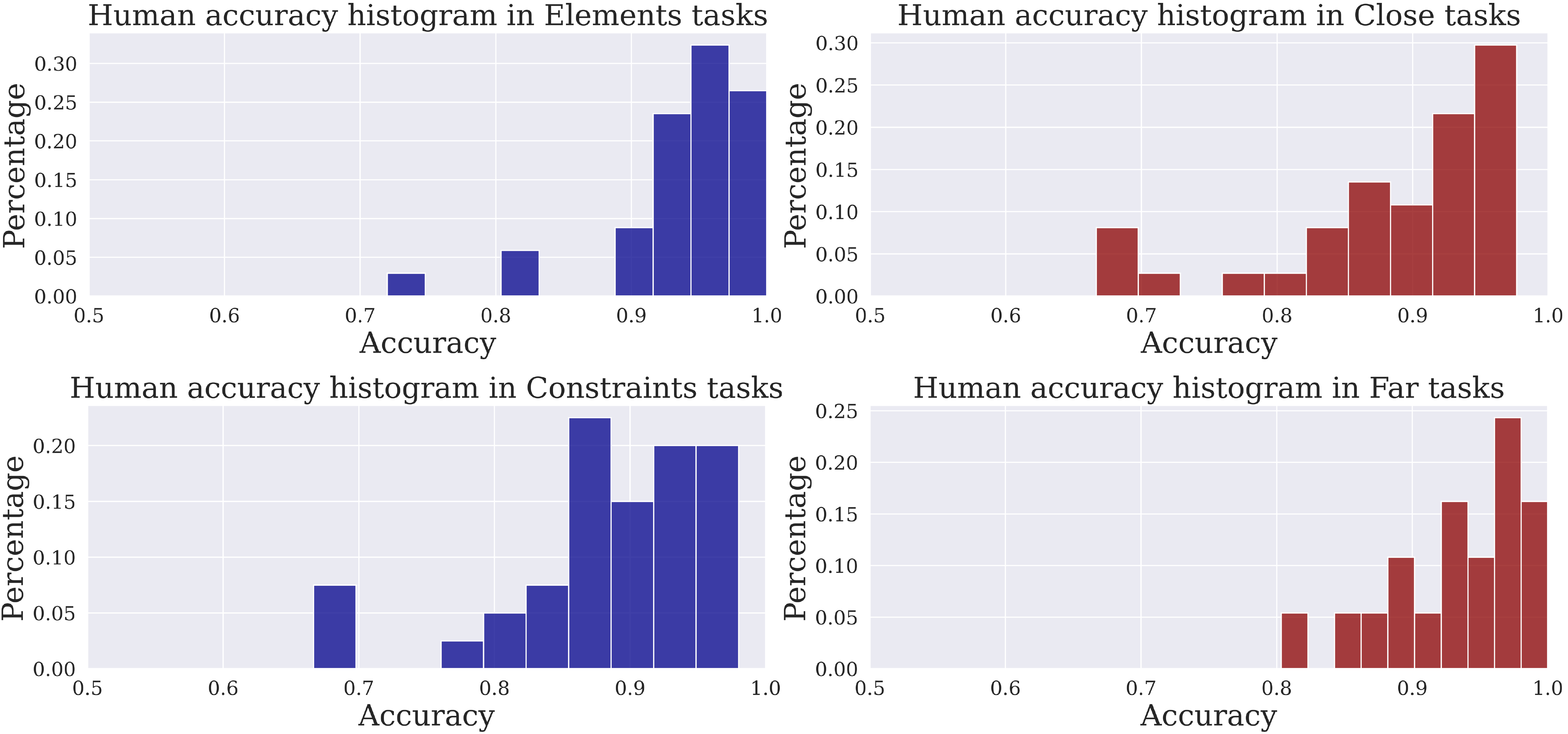}
  \end{center}
  \end{small}
  \caption{Histogram of human accuracies on the \elements and the \constraints datasets, as well as accuracies with \sdiff and \ldiff negative examples. The y-axis indicates the percentage of tasks with the specified accuracy on the x-axis.}
\label{fig:human_acc}
\end{wrapfigure}

We report task accuracy in Table~\ref{table:results}, scored as the percentage of participants correctly categorizing test images, averaged across all examples. We split the $15$ test images into two tasks, with the \sdiff task consisting of $5$ positive examples and $5$ negative examples from \sdiff, and the \ldiff task consisting of the same $5$ positive images with $5$ negative examples from \ldiff. Each task contains $10$ examples in total.
We see that human performance is strong across all tasks, with on-average higher scores in \ldiff compared to \sdiff, showing that the number of differences in construction rules affects the semantic distance between rendered realizations. Only tasks \texttt{LLL, CLCL}, and rhomboid yielded slightly better performance in \sdiff than \ldiff. Out of all tasks, \texttt{LLC, CCC, LLLC} with \sdiff negatives are more difficult for humans (with equally poor performance across all test images), which we hypothesize is due to more subtle constraint intersections. In general, humans perform well on this generalization task.  

We show accuracy histograms in Figure~\ref{fig:human_acc}, with the left two plots depicting results from concepts in \elements and \constraints, and the right two plots depicting results when calculated with \sdiff negative examples and \ldiff negative examples. Humans are more sensitive to concepts in \elements, which are complex constructions that test the exactness of shapes (\eg, squares and equilateral triangles), and slightly less sensitive to concepts in \constraints, which test the precise relationships between objects (\eg, constrained contact point between the end of a line and the center of a circle).

Participants could generate infinitely many rules consistent with the positive images seen in the few-shot examples (\eg, a ``wug'' can have its own prototype for each example, as there are no negative examples), and there is potential ambiguity as to which are the correct construction rules of the concept. Despite this wide range of possible generalization patterns, the generalization rule chosen by humans corresponds well to the Euclidean construction universe. 

\begin{table*}[htb!] 
\caption{Human accuracy across all $74$ tasks in \model.}
\begin{center}
\begin{small}
\begin{sc}
\begin{tabular}{l|ll||l|ll}
     concept & \sdiff & \ldiff & concept & \sdiff & \ldiff  \\
     \toprule
angle & 0.9767 & 0.9833 & lll & 0.9700 & 0.9667 \\
perp bisector & 0.9367 & 0.9833 & cll & 0.9467 & 0.9667  \\
ang bisector & 0.9433 & 0.9533 & llc & 0.6767 & 0.9233 \\
sixty ang & 0.8233 & 0.9533 & ccl & 0.8700 & 0.8833  \\
radii & 0.9233 & 0.9600 & lcc & 0.8867 & 0.9633 \\
diameter & 0.9567 & 1.0000 & ccc & 0.6667 & 0.8767  \\
segment & 0.9300 & 0.9833 & llll & 0.8833 & 0.9767  \\
rectilinear & 0.9000 & 0.9033 & lllc & 0.6667 & 0.8867  \\
triangle & 0.9633 & 0.9767 & clll & 0.8367 & 0.9033  \\
quadrilateral & 0.9167 & 0.9267 & clcl & 0.8700 & 0.8567  \\
eq t & 0.9533 & 0.9800 & llcc & 0.8867 & 0.9333 \\
right ang t & 0.7200 & 0.8133 & cccl & 0.9233 & 0.9333  \\
square & 0.8933 & 0.9867 & clcc & 0.8633 & 0.9000  \\
rhombus & 0.9367 & 0.9667 & cccc & 0.8167 & 0.8800  \\
oblong & 0.9666 & 0.9900 & tll & 0.9467 & 0.9800  \\
rhomboid & 0.9700 & 0.9300 & llt & 0.9267 & 0.9400  \\
parallel l & 0.9500 & 0.9567 & tcl & 0.9533 & 0.9633  \\
&  &  & clt & 0.9533 & 0.9633  \\
&  &  & tcc & 0.9533 & 0.9633  \\
&  &  & cct & 0.9533 & 0.9633  \\
    \bottomrule
  \end{tabular}
\end{sc}
\end{small}
\end{center}
\label{table:results}
\end{table*}

\subsection{Model Benchmarks}
\label{model_benchmarks}

\begin{wrapfigure}{r}{0.7\textwidth}
  \begin{small}
  \begin{center}
    \includegraphics[width=0.7\textwidth]{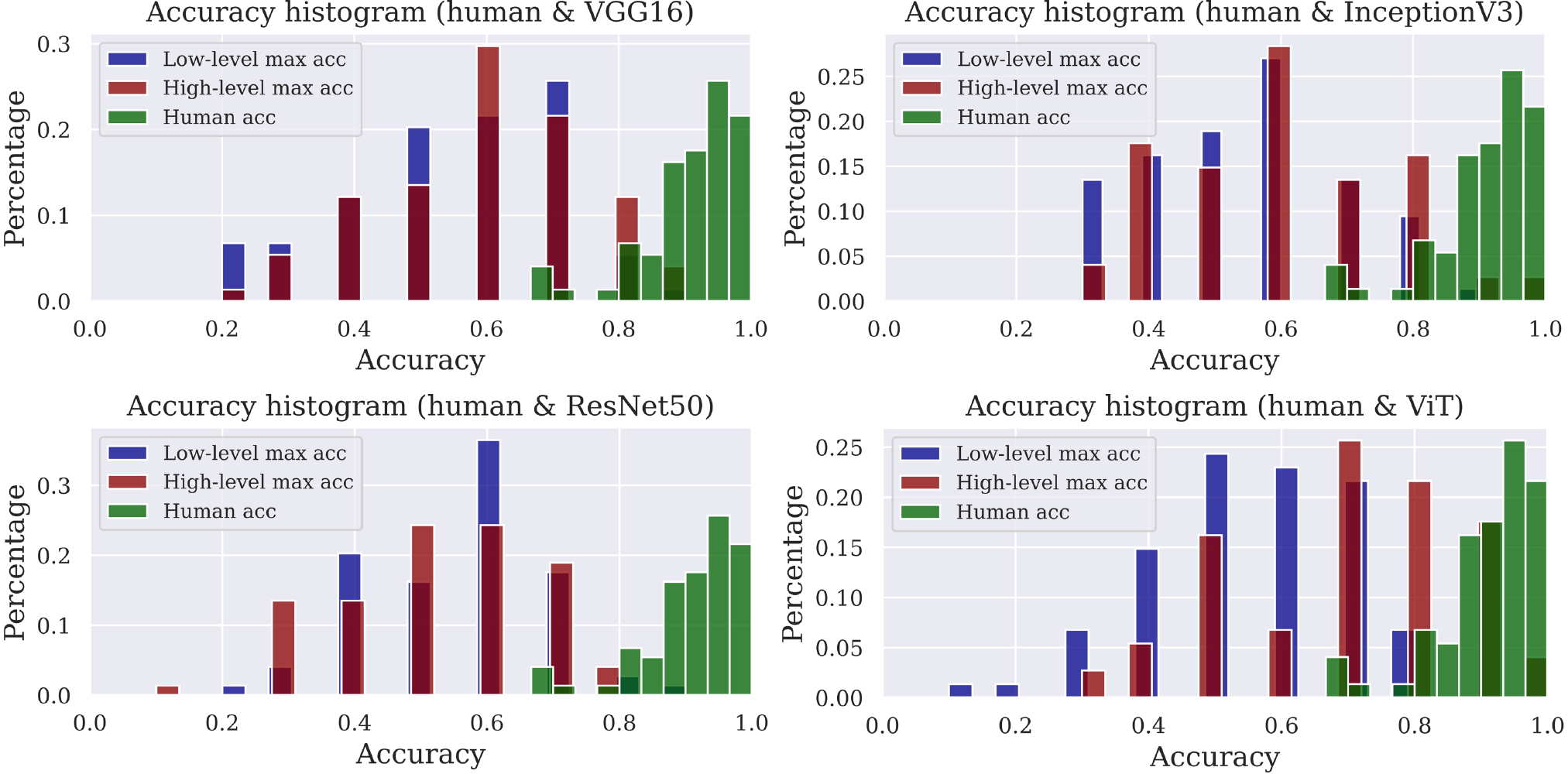}
  \end{center}
  \end{small}
  \caption{Histogram of human and maximum low-level and high-level feature accuracies of various vision models.} 
\label{fig:h_v_f}
\end{wrapfigure}

We benchmark pretrained vision models' performance on the \model task, to evaluate few-shot generalization (with no meta-learning or fine-tuning) in the Euclidean geometry universe. We measure performance of features from ImageNet-pretrained VGG16 \citep{simonyan2014very}, ResNet50 \citep{he2016deep}, InceptionV3 \citep{szegedy2016rethinking}, and Vision Transformer \citep{dosovitskiy2020image}, and evaluate both low-level features and high-level features for each of the models. Low-level features are outputs of earlier layers in neural networks, which tend to capture low-level information such as edges and primitive shapes, while high-level features are outputs of later layers that tend to capture more high-level semantic information \citep{zeiler2014visualizing}. We detail how we define the layers for each baseline in the Appendix. To evaluate these features, we extract features $\phi$ from the few-shot $t$ reference images of the target concept, to create a prototype target feature, $T = \frac{1}{n} \sum_{i=1}^{n}\phi(t_i)$. We classify a test image $r$ as in-concept if it is closer than a threshold to the prototype: $f|T - \phi(r)|<\theta$, where $f$ is the normalizing function between $0 \sim 1$.
The threshold is fit by selecting the best-performing normalized distance threshold across all 74 tasks, for given features. 
(By fitting this free threshold, we bias the reported accuracy in favor of models.)

\begin{table*}[htb!] 
\caption{Human accuracy compared to low and high-level feature accuracy of vision models across $37$ concepts in \model, each concept containing averaged accuracies between \sdiff and \ldiff tasks.}
\begin{center}
\begin{small}
\begin{sc}
\begin{tabular}{l|l||ll||ll||ll||ll}
     & Human & VGG16 & & RN50 & & InV3 & & ViT \\
     auc &  & \lfeat & \hfeat & \lfeat & \hfeat & \lfeat & \hfeat & \lfeat & \hfeat  \\
     \toprule
angle & 0.98 & 0.50 & 0.45 & 0.45 & 0.50 & 0.40 & 0.45 & 0.50 & 0.40 \\
perp bisector & 0.96 & 0.70 & 0.85 & 0.70 & 0.50 & 0.65 & 0.90 & 0.70 & 0.85 \\
ang bisector & 0.95 & 0.50 & 0.60 & 0.50 & 0.65 & 0.60 & 0.60 & 0.50 & 0.75 \\
sixty ang & 0.89 & 0.25 & 0.65 & 0.35 & 0.45 & 0.55 & 0.45 & 0.35 & 0.70 \\
radii & 0.94 & 0.75 & 0.70 & 0.75 & 0.60 & 0.65 & 0.80 & 0.75 & 0.75 \\

diameter & 0.98 & 0.30 & 0.55 & 0.40 & 0.40 & 0.50 & 0.75 & 0.45 & 0.85 \\
segment & 0.96 & 0.45 & 0.60 & 0.60 & 0.40 & 0.30 & 0.65 & 0.55 & 0.65 \\
rectilinear & 0.90 & 0.65 & 0.35 & 0.60 & 0.45 & 0.55 & 0.60 & 0.65 & 0.45 \\
triangle & 0.97 & 0.65 & 0.45 & 0.50 & 0.40 & 0.40 & 0.55 & 0.35 & 0.60 \\
quadrilateral & 0.92 & 0.50 & 0.70 & 0.60 & 0.50 & 0.55 & 0.75 & 0.60 & 0.60 \\
eq t & 0.97 & 0.40 & 0.85 & 0.50 & 0.70 & 0.65 & 0.65 & 0.50 & 0.55 \\

right ang t & 0.77 & 0.65 & 0.75 & 0.60 & 0.70 & 0.60 & 0.55 & 0.70 & 0.55 \\
square & 0.94 & 0.70 & 0.80 & 0.65 & 0.45 & 0.85 & 0.70 & 0.60 & 0.75 \\
rhombus & 0.95 & 0.70 & 0.55 & 0.60 & 0.55 & 0.60 & 0.55 & 0.70 & 0.60 \\
oblong & 0.98 & 0.45 & 0.55 & 0.45 & 0.45 & 0.50 & 0.70 & 0.45 & 0.70 \\
rhomboid & 0.95 & 0.70 & 0.60 & 0.65 & 0.55 & 0.50 & 0.50 & 0.65 & 0.75 \\
parallel l & 0.95 & 0.55 & 0.35 & 0.50 & 0.40 & 0.45 & 0.55 & 0.45 & 0.85 \\
\midrule
lll & 0.97 & 0.50 & 0.70 & 0.55 & 0.40 & 0.55 & 0.35 & 0.55 & 0.80 \\
cll & 0.96 & 0.25 & 0.60 & 0.40 & 0.30 & 0.30 & 0.60 & 0.30 & 0.60 \\
llc & 0.80 & 0.65 & 0.55 & 0.60 & 0.55 & 0.55 & 0.65 & 0.60 & 0.60 \\
ccl & 0.88 & 0.65 & 0.50 & 0.55 & 0.65 & 0.30 & 0.60 & 0.55 & 0.75 \\
lcc & 0.93 & 0.45 & 0.70 & 0.45 & 0.70 & 0.65 & 0.70 & 0.45 & 0.85 \\
ccc & 0.77 & 0.60 & 0.55 & 0.60 & 0.70 & 0.70 & 0.75 & 0.60 & 0.90 \\
llll & 0.93 & 0.60 & 0.30 & 0.50 & 0.30 & 0.30 & 0.50 & 0.50 & 0.50 \\
lllc & 0.78 & 0.30 & 0.65 & 0.40 & 0.55 & 0.50 & 0.55 & 0.30 & 0.80 \\
clll & 0.87 & 0.55 & 0.65 & 0.60 & 0.65 & 0.65 & 0.50 & 0.70 & 0.65 \\
clcl & 0.86 & 0.60 & 0.65 & 0.65 & 0.45 & 0.40 & 0.35 & 0.60 & 0.85 \\
llcc & 0.91 & 0.75 & 0.65 & 0.85 & 0.55 & 0.80 & 0.70 & 0.80 & 0.70 \\
cccl & 0.93 & 0.35 & 0.60 & 0.50 & 0.65 & 0.35 & 0.65 & 0.45 & 0.75 \\
clcc & 0.88 & 0.60 & 0.55 & 0.55 & 0.60 & 0.60 & 0.75 & 0.55 & 0.80 \\
cccc & 0.85 & 0.80 & 0.65 & 0.70 & 0.60 & 0.75 & 0.50 & 0.75 & 0.95 \\
tll & 0.96 & 0.70 & 0.65 & 0.65 & 0.60 & 0.60 & 0.55 & 0.75 & 0.85 \\
llt & 0.93 & 0.55 & 0.40 & 0.55 & 0.65 & 0.55 & 0.65 & 0.50 & 0.85 \\
tcl & 0.96 & 0.65 & 0.70 & 0.70 & 0.55 & 0.60 & 0.60 & 0.65 & 0.45 \\
clt & 0.88 & 0.50 & 0.60 & 0.45 & 0.50 & 0.55 & 0.65 & 0.45 & 0.70 \\
tcc & 0.84 & 0.30 & 0.45 & 0.35 & 0.35 & 0.55 & 0.45 & 0.30 & 0.45 \\
cct & 0.79 & 0.55 & 0.70 & 0.45 & 0.60 & 0.60 & 0.55 & 0.50 & 0.85 \\
\midrule
\textbf{average} & \textbf{0.91} & \textbf{0.55} & \textbf{0.60} & \textbf{0.55} & \textbf{0.53} & \textbf{0.54} & \textbf{0.60} & \textbf{0.55} & \textbf{0.70} \\
    \bottomrule
  \end{tabular}
\end{sc}
\end{small}
\end{center}
\label{table:auc_results}
\end{table*}

In Table~\ref{table:auc_results}, we present accuracy of ImageNet-pretrained low-level and high-level features across different models. 
Humans substantially outperform features from vision models, showcasing the gap between human and model capabilities in Euclidean geometry concept learning.

We see that, on average across tasks, pretrained vision models perform poorly compared to humans. In general, high-level features perform slightly better than low-level features, though there are cases where this differs.  Interestingly, the high-level features from the Vision Transformer (ViT) outperform its convolutional network counterparts, and achieve human-level performance in some tasks. There are a few tasks where the visual feature accuracy outperforms humans, notably tasks \texttt{CCC, LLLC, CCCC, CCT}, where high-level features from ViT perform strongly. We hypothesize that this is because humans are not as sensitive to details of intersections between circles.

In Figure~\ref{fig:h_v_f}, we present histograms comparing maximum low and high-level feature accuracy to human accuracy, illustrating the gap in performance. In Table~\ref{table:correlation}, we report the Pearson correlation coefficient between the answers of humans and models. We see that humans are generally more aligned with high-level features than low-level ones. Additionally, though ViT achieves high accuracy, it is not correlated to human performance, indicating failure to generalize in the most human-like way.

\begin{table*}[htb!] 
\caption{Pearson's correlation to human accuracy across all $74$ tasks in Geoclidean.}
\begin{center}
\begin{small}
\begin{sc}
\begin{tabular}{l|l||l|l||l|l||l|l}
      VGG16 \lfeat & \hfeat & RN50 \lfeat & \hfeat & InV3 \lfeat & \hfeat & ViT \lfeat & \hfeat  \\
     \toprule
       -0.08947 & 0.0943 & 0.0341 & -0.1522 & -0.0820 & 0.2575 & -0.0056 & -0.0259 \\
    \bottomrule
  \end{tabular}
\end{sc}
\end{small}
\end{center}
\label{table:correlation}
\end{table*}

We report additional comparisons of low-level and high-level visual features from ImageNet-pretrained VGG16, ResNet50, InceptionV3, and Vision Transformer in the Appendix, and show that similar trends follow across data splits and performance metrics. We also include low-level and high-level feature visualizations in the Appendix, comparing Geoclidean tasks that require reasoning, to perception tasks involving simple geometric primitives and perturbations. These comparisons highlight Geoclidean as a unique and interesting test for vision models.

\section{Related Work}
\paragraph{Geometric reasoning datasets.} Prior geometric datasets generally fall into two main categories---with geometric objects for computer-aided design (CAD) and for plane geometry. In the first category, the SketchGraphs dataset models relational geometry in CAD design \citep{seff2020sketchgraphs}, and the ABC-Dataset includes parametric representations of 3D CAD models \citep{koch2019abc}. In the latter category, CSGNet presented a generated dataset of constructive solid geometry based on 2D and 3D synthetic programs with squares, circles, and triangles \citep{sharma2018csgnet}, while \cite{ellis2018learning} connected high–level Latex graphics programs with 2D geometric drawings. Works such as \cite{zhang2022plane, lu2021inter} proposed using datasets with annotated geometric primitives and relationships such as containment from geometry diagrams in textbooks. Others introduced reasoning benchmarks with geometric shapes, including Raven's progressive matrices \citep{matzen2010recreating, wang2015automatic, barrett2018measuring, zhang2019raven}, Bongard problems \citep{depeweg2018solving, nie2020bongard}, odd-one-out tasks \citep{mandziuk2019deepiq}, and a variety of reasoning challenges \citep{hill2019learning, zhao2021learning, el20202d, zhang2020machine}. Our work is more related to the latter of geometric shapes, and \model differs by targeting Euclidean geometry concept learning whose construction language 1) does not require specific coordinates, and 2) focuses on the construction steps that form semantically-complex geometric concepts and the constraints between geometric primitives that humans are intrinsically sensitive to.

\paragraph{Few-shot concept learning.}
Few-shot learning tasks range in complexity on both the input and task description axis. In the natural language processing domain, tasks such as FewRel \citep{han2018fewrel} and Few-NERD \citep{ding2021few} have been proposed for few-shot relation classification and entity recognition. \cite{goodman2008rational} introduced concept learning tasks with sequences generated from specified logical rules. In the vision domain, which we are interested in, commonly used tasks include those from \cite{lake2015human}, which introduced Omniglot as a collection of simple visual concepts collected from 50 writing systems, and from \cite{vinyals2016matching}, which proposed miniImageNet \citep{deng2009imagenet}, both for the task of one-shot classification. Works in other vision domains include \cite{massiceti2021orbit}, which explores the few-shot video recognition challenge, and \cite{xiao2020multi, gehler2008bayesian}, which examines the few-shot color constancy problem. \cite{triantafillou2019meta} created the meta-dataset as a diverse dataset for few-shot learning, with multiple tasks for meta-training such as \cite{maji2013fine, wah2011caltech, cimpoi2014describing, nilsback2008automated, houben2013detection, lin2014microsoft}. In comparison, we propose \model as a zero-shot meta-trained, few-shot generalization task that consists of labeled image renderings from a single target concept. 

\vspace{-0.2cm}
\section{Discussion}
An important contribution of our task is that it allows for better testing of vision models that aim to incorporate reasoning and high-level semantics. Additionally, Geoclidean's zero-shot meta-trained evaluation is especially significant, as many downstream tasks that may leverage pretrained models would greatly benefit from geometric reasoning, such as construction (LegoTron \citep{walsmanlegotron}, Physical Construction Tasks \cite{bapst2019structured}), physical reasoning (CLEVRER \citep{yi2019clevrer}, ThreeDWorld \citep{gan2020threedworld}), and shape understanding tasks (PartNet \citep{mo2019partnet}, ShapeNet \citep{chang2015shapenet}). Furthermore, numerous additional evaluation tasks can be built with the Geoclidean DSL and rendering library, such as those involving natural language or generated large-scale datasets. We include further analyses and discussion in the Appendix.

\section{Conclusion}
\label{conclusion}
We have introduced \model, a domain-specific language for the realization of the Euclidean geometry universe, and presented two datasets of few-shot concept learning to test generalization capability in the geometry domain. Humans considerably outperform vision models on \model tasks, and we believe that this gap illustrates the potential for improvement in learning visual features that align with human sensitivities to geometry. \model is thus an important generalization task that vision models are not yet sensitive to, and an effective benchmark for geometric concept learning.
Furthermore, such explorations of geometric generalization may help us to understand how human vision made the leap from natural forms to the Platonic forms so prevalent in modern design and engineering.




We expect minimal negative societal impact from the release of \model. 
We hope future work can build on the foundations of \model for augmenting vision models in areas such as geometric reasoning and construction, as well as in applications such as education, where geometry is both an essential academic subject and an introduction to proof-based mathematics. 



\paragraph{Acknowledgements}
We thank Gabriel Poesia and Stephen Tian for providing valuable feedback on the paper. This work is in part supported by the Stanford Institute for Human-Centered Artificial Intelligence (HAI), Center for Integrated Facility Engineering (CIFE), Analog, Autodesk, IBM, JPMC, Salesforce, and Samsung. JH is supported by the Knight Hennessy fellowship and the NSF Graduate Research Fellowship.


{
\small
\bibliographystyle{plainnat}
\bibliography{reference}
}

\newpage

\section*{Checklist}

\begin{enumerate}

\item For all authors...
\begin{enumerate}
  \item Do the main claims made in the abstract and introduction accurately reflect the paper's contributions and scope?
    \answerYes{}
  \item Did you describe the limitations of your work?
    \answerYes{}, see Section~\ref{human_performance}.
  \item Did you discuss any potential negative societal impacts of your work?
    \answerYes{}, see Section~\ref{conclusion}.
  \item Have you read the ethics review guidelines and ensured that your paper conforms to them?
    \answerYes{}
\end{enumerate}

\item If you are including theoretical results...
\begin{enumerate}
  \item Did you state the full set of assumptions of all theoretical results?
    \answerNA{}
	\item Did you include complete proofs of all theoretical results?
    \answerNA{}
\end{enumerate}

\item If you ran experiments (e.g. for benchmarks)...
\begin{enumerate}
  \item Did you include the code, data, and instructions needed to reproduce the main experimental results (either in the supplemental material or as a URL)?
    \answerYes{}, see Appendix.
  \item Did you specify all the training details (e.g., data splits, hyperparameters, how they were chosen)?
    \answerNA{There is no training involved.}
	\item Did you report error bars (e.g., with respect to the random seed after running experiments multiple times)?
    \answerNA{There is no training involved.}
	\item Did you include the total amount of compute and the type of resources used (e.g., type of GPUs, internal cluster, or cloud provider)?
    \answerYes{}, see Appendix.
\end{enumerate}

\item If you are using existing assets (e.g., code, data, models) or curating/releasing new assets...
\begin{enumerate}
  \item If your work uses existing assets, did you cite the creators?
    \answerNA{}
  \item Did you mention the license of the assets?
    \answerYes{}, see Appendix.
  \item Did you include any new assets either in the supplemental material or as a URL?
    \answerYes{}, see Appendix.
  \item Did you discuss whether and how consent was obtained from people whose data you're using/curating?
    \answerYes{}, see Appendix.
  \item Did you discuss whether the data you are using/curating contains personally identifiable information or offensive content?
    \answerYes{}, see Appendix.
\end{enumerate}

\item If you used crowdsourcing or conducted research with human subjects...
\begin{enumerate}
  \item Did you include the full text of instructions given to participants and screenshots, if applicable?
    \answerYes{}, see Appendix.
  \item Did you describe any potential participant risks, with links to Institutional Review Board (IRB) approvals, if applicable?
    \answerYes{}, see Appendix.
  \item Did you include the estimated hourly wage paid to participants and the total amount spent on participant compensation?
    \answerYes{}, see Appendix.
\end{enumerate}

\end{enumerate}


\end{document}